\documentclass[letterpaper, 10 pt, conference]{ieeeconf} 
\IEEEoverridecommandlockouts
\usepackage{times} 
\usepackage{amsmath} 
\usepackage{amssymb}  
\usepackage{booktabs}
\usepackage{physics}

\usepackage{graphicx}
\usepackage{url}
\usepackage{float}
\usepackage{dblfloatfix}
\usepackage{subfig}
\usepackage{textcomp}

\usepackage{gensymb}
\usepackage{multicol,tabularx,capt-of}
\usepackage{multirow}

\usepackage{booktabs}
\usepackage{multirow}

\usepackage{algorithm}
\usepackage{algpseudocode}

\usepackage{listings}
\usepackage{xcolor}
\usepackage{hyperref}

\title{\LARGE \bf
RoboLLM: Robotic Vision Tasks Grounded on Multimodal Large Language Models
}

\author{Zijun Long$^{1}$, George Killick$^{1}$, Richard McCreadie$^{1}$ and Gerardo Aragon-Camarasa$^{1}$
\thanks{This research has been supported by EPSRC Grant No. EP/S019472/1}
\thanks{$^{1}$ School of Computing Science, University of Glasgow, G12 8QQ, Scotland, United Kingdom {\tt\small z.long.2@research.gla.ac.uk}}%
}

\begin{document}

\maketitle
\thispagestyle{empty}
\pagestyle{empty}

\begin{abstract}

Robotic vision applications often necessitate a wide range of visual perception tasks, such as object detection, segmentation, and identification. While there have been substantial advances in these individual tasks, integrating specialized models into a unified vision pipeline presents significant engineering challenges and costs. Recently, Multimodal Large Language Models (MLLMs) have emerged as novel backbones for various downstream tasks. We argue that leveraging the pre-training capabilities of MLLMs enables the creation of a simplified framework, thus mitigating the need for task-specific encoders. Specifically, the large-scale pretrained knowledge in MLLMs allows for easier fine-tuning to downstream robotic vision tasks and yields superior performance. We introduce the RoboLLM framework, equipped with a BEiT-3 backbone, to address all visual perception tasks in the ARMBench challenge—a large-scale robotic manipulation dataset about real-world warehouse scenarios. RoboLLM not only outperforms existing baselines but also substantially reduces the engineering burden associated with model selection and tuning. All the code used in this paper can be found in \url{https://github.com/longkukuhi/armbench}.


\end{abstract}

\section{INTRODUCTION}

Recent advances in deep learning have led to the emergence of Multimodal Large Language Models (MLLMs) such as BLIP2~\cite{DBLP:conf/icml/0008LSH23} and BEiT-3~\cite{beit3}. These MLLMs, trained on large-scale datasets, act as general-purpose encoders for multiple modalities and offer substantial potential for transfer to various applications~\cite{RN220,RN155,RN150,RN153}. While they have shown state-of-the-art performance in fields such as conversational agents, their utility in robotic vision remains largely unexplored. Early studies, like PaLM-E~\cite{driess2023palm}, have begun to delve into this area but primarily focus on vision and language understanding, largely overlooking the unique real-world challenges of robotic vision, such as task-specific camera poses, variable lighting, and clutter. Therefore, prompted by the success of MLLMs in other domains, this paper addresses the question: \textit{To what extent can MLLMs improve vision tasks specific to the robotics domain?}

The recently introduced ARMBench dataset~\cite{armbench} from Amazon exemplifies the complexities inherent to robotic vision. It comprises three critical robotic vision perception tasks: object instance segmentation, object identification, and defect detection, and presents a large-scale representation of real-world scenarios—specifically, object grasping and manipulation tasks in Amazon warehouses. These tasks demand a vision system capable of handling an extremely large range of objects (e.g., 190K+ unique objects), robust to variable lighting conditions, and capable of operating effectively in cluttered environments. Moreover, the ever-changing inventory of warehouses requires the vision system to perform well in transfer learning scenarios. The individual tasks, though extensively researched, have yielded task-specific models, complicating their integration into a unified vision pipeline. Such integration involves selecting the right model for each task, coordinating them effectively, and fine-tuning each one -- a process that is both time-consuming and challenging in terms of engineering.

We argue that MLLMs have better transfer capability than previous task-specific models, owing to MLLMs' large-scale pre-training, which allows them to serve as robust backbones for these tasks. This substantially reduces the engineering complexity of developing different backbones for multiple tasks and the time required for model selection and tuning. Additionally, MLLMs have shown particular resilience to out-of-distribution examples, such as objects presented in novel poses or amidst visual distractions, which are critical issues in robotic vision perception tasks.
In this work, we introduce RoboLLM, a framework designed to adapt MLLMs to the multifaceted challenges of robotic vision, using the ARMBench dataset for evaluation. 
Our main contributions are as follows:
\begin{itemize}
\vspace{-0.5mm}
\item We present RoboLLM, a generalized framework that employs pre-trained MLLMs as backbones for tackling complex robotic vision tasks, including object instance segmentation, object identification, and defect detection.
\item We are the first work to address all three key vision tasks in the ARMBench dataset, representative of challenging, large-scale robotic manipulation scenarios.
\item We introduce a lightweight variant of the BEiT-3 architecture aimed at increased performance and efficiency, thus broadening its applicability to resource-constrained robotic applications.
\item Our experiments show that our RoboLLM achieves state-of-the-art results across all three ARMBench tasks.
\item We further demonstrate RoboLLM's robustness to object number variance and its superior performance on out-of-distribution examples in the object segmentation task, where previous works fail. More importantly, RoboLLM solves the object identification task with a 97.8\% recall@1.
\end{itemize}

\section{BACKGROUND}

\subsection{Emergence of Multimodal Large Language Models}

Recently, Multimodal Large Language Models (MLLMs) have achieved new state-of-the-art result, delivering superior performance across a wide range of vision and vision-language benchmarks. Their exceptional capabilities are most notable in few-shot and zero-shot scenarios, as evidenced by a series of studies~\cite{DBLP:conf/nips/AlayracDLMBHLMM22,DBLP:conf/icml/0008LSH23,beit3,driess2023palm,DBLP:journals/tmlr/YuWVYSW22}. Their efficacy comes from extensive pretraining on large-scale corpora of image-text data, enabling them with superior transfer capabilities. Palm-E, a pioneer in the use of MLLMs in the robotics domain, has exhibited state-of-the-art performance in embodied robotic planning scenarios~\cite{driess2023palm}. Open X-Embodiment \cite{RN148} amalgamates extensive datasets, showcasing the efficacy of transfer learning with large-scale models. This approach significantly enhances the capabilities of various robots by leveraging shared knowledge across different platforms. Furthermore, research indicates that the multi-modal large-scale training of image-text pairs results in models with enhanced robustness to out-of-distribution examples~\cite{mayo2023how} compared to their unimodal counterparts. This let MLLMs show superior performance in unimodal tasks, such as vision tasks, when compared to vision-specific pretrained models~\cite{beit3, DBLP:conf/icml/0008LSH23}. As such, MLLMs present a compelling case for serving as backbone encoders in complex robotic vision applications, which often consist of multiple sub-tasks. This enables us to improve the effectiveness of robotic vision pipeline yet significantly reduce the engineering efforts and tuning time.

\subsection{Image Segmentation}
Object instance segmentation involves simultaneously predicting pixel-level instance masks and their corresponding class labels. Though the most prevalent backbones for these detector models have been Convolutional Neural Networks (ConvNets)~\cite{convet}, such as R-CNN~\cite{rcnn} and Faster R-CNN~\cite{DBLP:journals/corr/RenHG015}, the Vision Transformer (ViT)~\cite{vit} has emerged as a potent alternative for image classification tasks. The original ViT architecture is non-hierarchical; this characteristic hinders its applicability in object instance segmentation due to a lack of innate translational equivariance~\cite{DBLP:journals/corr/shifteq} and difficulties in handling high-resolution inputs because of the quadratic complexity of self-attention. Therefore, some models aim to mitigate these challenges by incorporating ConvNet designs into ViT, integrating hierarchical structures and translation-equivariant priors such as convolutions, pooling, and sliding windows (e.g., Swin~\cite{swin}, MViT~\cite{mvit}, PVT~\cite{pvt}). This approach compromises the model's general applicability by coupling pre-training and fine-tuning requirements. Subsequent efforts have explored plain ViT backbones specifically for object instance segmentation, such as UViT~\cite{uvit}, a single-scale Transformer for object detection. Unlike UViT, ViTDet~\cite{vitdet} offers an approach that retains the task-agnostic nature of ViT backbones, thereby facilitating their broader applicability. In this study, we extend the line of inquiry initiated by ViTDet in the field of robotic vision. Our approach employs a plain backbone architecture decoupled from the detection task, enabling the reuse of the same backbone across various robotic vision perception tasks. This strategy significantly alleviates the engineering challenges related to model selection and fine-tuning.

\subsection{Image Retrieval}

The object identification task in ARMBench can be tackled as an image retrieval task~\cite{ge2021structured,ge2023cross}, a well-studied area in computer vision with applications in robotics for scene localization~\cite{anoosheh2019night} and place recognition~\cite{DBLP:conf/iros/ChenMSC17,RN151}. Traditional methods have focused on aggregating ConvNet feature maps using various pooling techniques such as R-MAC~\cite{DBLP:journals/corr/ToliasSJ15} and GeM~\cite{DBLP:journals/pami/RadenovicTC19}, the latter of which has achieved state-of-the-art performance. Recently, the advent of vision transformers~\cite{vit} has initiated a shift away from ConvNets, often surpassing them in performance~\cite{irtransformer, boostingtrans} and reducing the need for specialized aggregation methods~\cite{irtransformer}. As for the objective functions for training, they have commonly employed contrastive or triplet loss functions~\cite{chopra2005learning, melekhov2016siamese, gordo2016deep,RN157,RN156,long2024clce}. Thus, contrastive learning has been shown to be effective in training image retrieval tasks.

In this work, we fine-tune a Multimodal Large Language Model (BEiT-3) with contrastive loss, leveraging its large-scale pretraining for effective object identification in the context of robotic vision. We argue that if the produced embeddings are sufficiently good, there is no need for complex feature processing methods or ranking techniques. This aligns well with the objective of RoboLLM, which aims to reduce engineering efforts in model tuning and enhance efficiency.

\subsection{Defect Detection}
Research in defect detection has been primarily oriented toward identifying surface defects in materials such as fabric, metals, and concrete~\cite{DBLP:journals/access/CaoLH20, DBLP:journals/tim/LuoFLYS20, rasheed2020fabric}. Though the primary goal is to ascertain the existence of a defect, certain applications necessitate the specific type of defect to be classified, localized, and segmented~\cite{tabernik2020segmentation, li2021defectnet}. Feng et al.~\cite{feng2019using} employed an autoencoder pretraining method for defect detection with limited training data, while Hu et al.~\cite{DBLP:journals/tip/HuGWRJYY21} introduced a lightweight spatial-temporal model incorporating local attention and PCA reduction to detect thermography defects. In contrast, ARMBench presents a large-scale challenge in which defects can manifest in various forms. Obtaining examples of defects for all objects is often infeasible, necessitating a model that can generalize to defects in unseen forms. We hypothesize that MLLMs, pretrained on large-scale datasets, can significantly improve the model's ability to handle out-of-distribution samples.

\section{MATERIALS and METHODS}

\begin{figure*}
    \centering
    \includegraphics[width=0.8\textwidth]{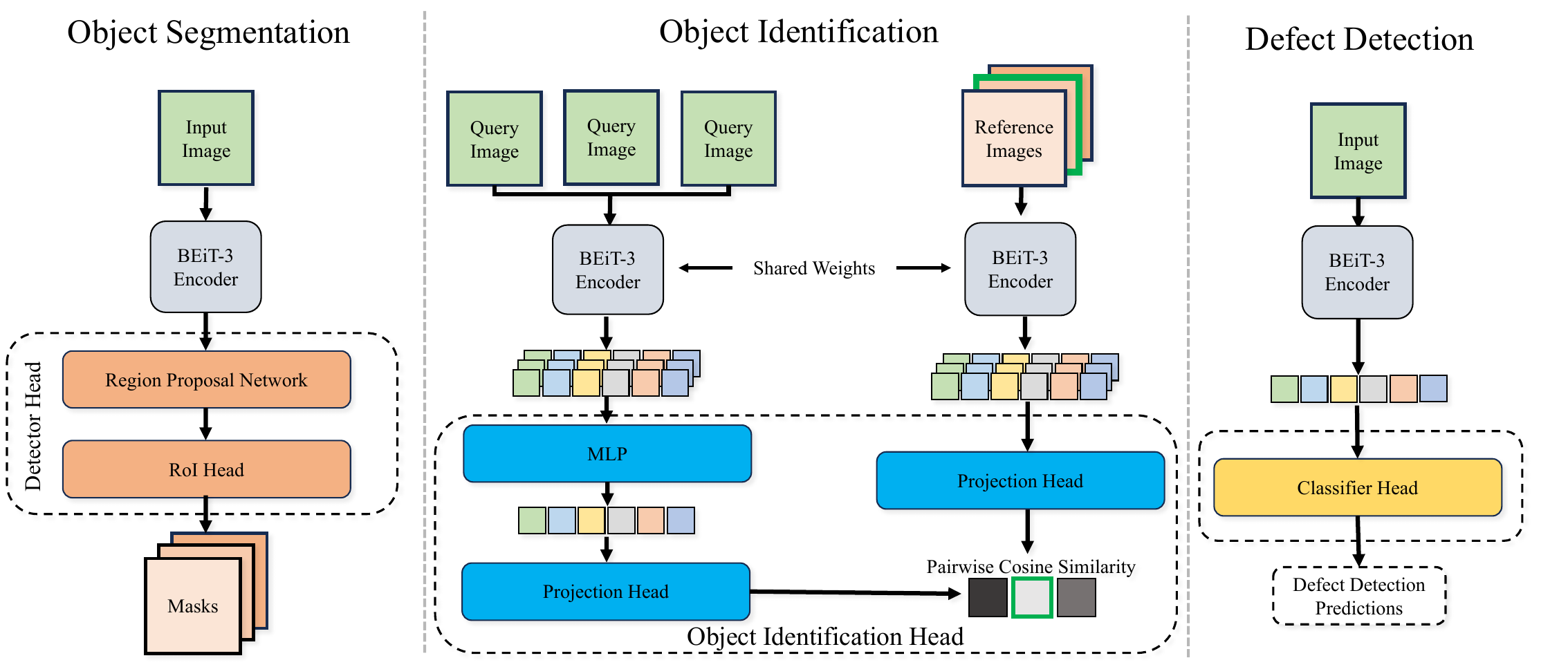}
    \caption{The task-specific adaptions to BEiT-3, proposed in our RoboLLM framework, for tackling each challenge in ARMBench. BEiT-3's large-scale vision-language pretraining allows it to be easily and effectively transferred to downstream tasks.}
    \label{fig:framework}
    \vspace{-3mm}
\end{figure*}

\subsection{Overview}

The proposed framework, RoboLLM, addresses three primary vision tasks in robotic manipulations as defined in the ARMBench dataset: \textit{object segmentation, object identification, and defect detection}. As illustrated in Figure~\ref{fig:framework}, the architecture of RoboLLM is inherently modular. This modular design allows for integrating a variety of Multimodal Large Language Models (MLLMs) as backbone encoders alongside task-specific heads tailored for distinct robotic vision tasks. This design principle of decoupling the backbone from specific downstream tasks offers several advantages, including ease of maintenance and flexibility for quick and easy adaptation to different vision tasks while fully exploiting the benefits of large-scale pre-trained models. Task-specific heads are integrated to tackle each unique vision challenge, with the choice of heads justified by their proven effectiveness in the respective tasks. Subsequent sections elaborate on the role of the backbone encoder and provide specific configurations for each vision task in the ARMBench dataset.

\subsection{Backbone Encoder}

\begin{figure}
    \centering
    \includegraphics[width=7cm]{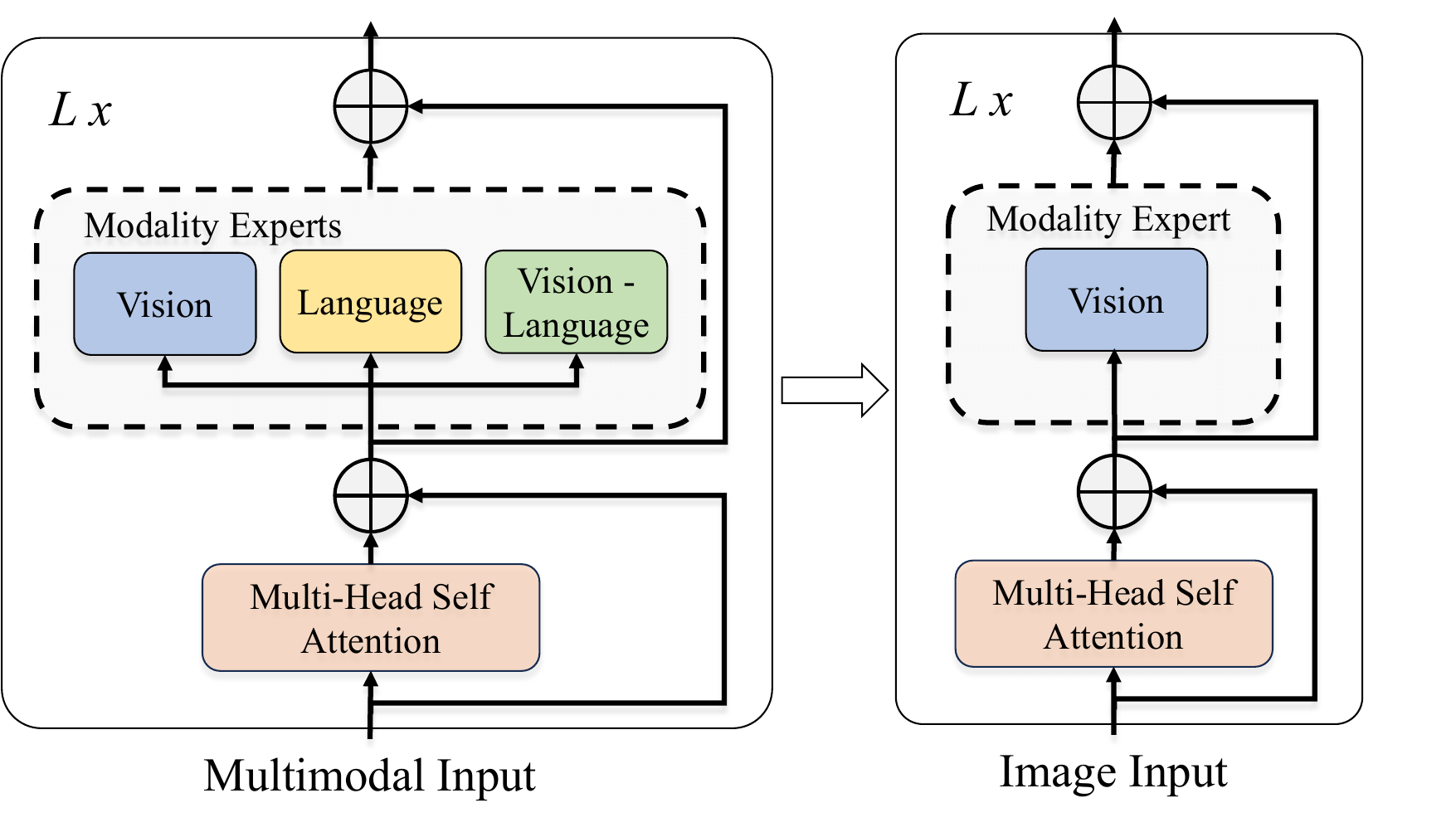}
    \caption{Our lightweight modification of the BEiT-3 backbone only remains the vision experts. }
    \label{fig:backbone_simplication}
    \vspace{-4mm}
\end{figure}

\looseness -1 The backbone encoder is a critical component in the RoboLLM framework, which is responsible for producing feature maps for downstream tasks. It is based on a widely-used MLLM architecture featuring a MultiWay Transformer design~\cite{beit3}. Our design philosophy allows the backbone to be replaced with future MLLM developments, making the system future-proof in the rapidly evolving domain of robotic vision. Currently, we believe BEiT-3 is the most suitable option owing to its state-of-the-art performance in various vision benchmarks~\cite{beit3}. Moreover, BEiT-3 demonstrates that features learned in a multi-modal setting are more task-agnostic and generalized, thereby offering better transferability to different vision tasks without the need for extensive fine-tuning. These characteristics of BEiT-3 make it highly suitable for robotic vision tasks.

\looseness -1 The MultiWay Transformer blocks, illustrated on the left side of Figure~\ref{fig:backbone_simplication}, comprise a shared self-attention module and a pool of feed-forward networks known as modality experts. These experts are designed for handling different types of data. However, this paper is focused only on addressing robotic vision perception tasks, thereby eliminating the need for other modality experts besides vision. Thus, our streamlined architecture, as shown on the right side of Figure~\ref{fig:backbone_simplication}, eliminates non-vision modality experts, reducing computational overhead and the parameter count from 222 million to 87 million in the BEiT-3 Base model. This design choice not only improves computational efficiency but also retains the flexibility to incorporate other modalities in future iterations of the task. Importantly, the feature maps generated by the backbone are exploited across all three vision tasks addressed in this work, demonstrating the framework's effectiveness and efficiency.

\subsection{Object Segmentation}
Object segmentation is the prerequisite task within the ARMBench dataset, aiming to identify and delineate individual objects in containers. Aligned with RoboLLM's overarching philosophy of decoupling the backbone from downstream tasks, we employ a plain-backbone approach for object segmentation. Unlike most object detection approaches, this plain-backbone detection approach does not require any pretraining on detection tasks, thereby eliminating hierarchical constraints on the backbone. This is a necessary step, given the absence of a segmentation task during MLLM pretraining. Specifically, we transfer MLLM to object segmentation tasks, with modifications performed exclusively during the fine-tuning stage. The rationale behind this choice is to maximize the benefits accrued from pre-training on large-scale datasets, which in turn improves segmentation performance. Thus, to translate the capabilities of MLLM into practical applications for object segmentation, a task-specific head is crucial for integrating and adapting the backbone's generic feature maps to the task at hand.

\subsubsection{Task-Specific Head}
The task-specific head for object segmentation integrates a detector based on a Cascade Mask R-CNN~\cite{cai2019cascade}. This modular design facilitates seamless interaction with the feature maps generated by the backbone and is extensible for compatibility with other detector heads.

\subsubsection{Feature Pyramid Construction}
The cornerstone of our object detection strategy is the construction of a feature pyramid, following~\cite{DBLP:conf/eccv/LiMGH22}. For this, we utilize only the last embeddings from the backbone, which is hypothesized to contain the most discriminative features. We then execute a series of parallel convolutions and deconvolutions to generate multi-scale feature maps. Specifically, starting with the default Vision Transformer (ViT) embeddings with a scale of \(\frac{1}{16}\), we produce feature maps at scales of \(\frac{1}{32}, \frac{1}{16}, \frac{1}{8}, \frac{1}{4}\) via convolutional strides of \(2, 1, \frac{1}{2}, \frac{1}{4}\), respectively.

\subsubsection{Object Detection Procedure}

These generated multi-scale feature maps are processed through the Cascade Mask R-CNN detector head. A Region Proposal Network (RPN) proposes candidate object bounding boxes, followed by mask generation for each object via a Region-of-Interest (ROI) head, which extracts pertinent features from the RPN.

\subsubsection{Advantages and Rationale}
Our design offers multiple benefits. Firstly, the plain-backbone approach necessitates only modest modifications during the fine-tuning stage, which preserves the generality and extensibility of the pre-trained MLLMs. Secondly, the task-agnostic nature of our backbone allows for seamless interchangeability with various detector heads, providing flexibility in addressing a broad range of object segmentation tasks. Lastly, the selective use of only the last feature map for object detection aims to utilize the most potent features, thereby enhancing the system's overall efficacy and efficiency.

\definecolor{codegreen}{rgb}{0,0.6,0}
\definecolor{codegray}{rgb}{0.5,0.5,0.5}
\definecolor{codepurple}{rgb}{0.58,0,0.82}
\definecolor{backcolour}{rgb}{0.95,0.95,0.92}

\lstdefinestyle{mystyle}{
    backgroundcolor=\color{backcolour},   
    commentstyle=\color{codegreen},
    keywordstyle=\color{magenta},
    numberstyle=\tiny\color{codegray},
    stringstyle=\color{codepurple},
    basicstyle=\ttfamily\footnotesize,
    breakatwhitespace=false,         
    breaklines=true,                 
    captionpos=b,                    
    keepspaces=true,                 
    numbers=left,                    
    numbersep=5pt,                  
    showspaces=false,                
    showstringspaces=false,
    showtabs=false,                  
    tabsize=2
}

\lstset{style=mystyle}

\begin{figure}
\begin{lstlisting}[language=Python, escapeinside={(*}{*)}]
import numpy as np
# Q[n, h, w, c] - minibatch of query images 
# R[n, h, w, c] - minibatch of reference images
# (*\textcolor{codegreen}{W\textsubscript{i}}*)[(*\textcolor{codegreen}{d\textsubscript{i}}*), (*\textcolor{codegreen}{d\textsubscript{e}}*)] - learned proj of image to embedding
# labels - labels of whether Query image and 
# reference image are the same class

# extract feature representations of each modality
if using both pre-pick and post-pick images:
    # BEiT-3 Encoder
    (*\texorpdfstring{Q\textsubscript{f}}{like this}*) = encoder(Q) #[n, 3 x (*\textcolor{codegreen}{d\textsubscript{i}}*)]
    # Feature Aggregation 
    (*\texorpdfstring{Q\textsubscript{f}}{like this}*) = MLP((*\texorpdfstring{Q\textsubscript{f}}{like this}*)) #[n, 3 x (*\textcolor{codegreen}{d\textsubscript{i}}*)] to [n, (*\textcolor{codegreen}{d\textsubscript{i}}*)]
else:
    (*\texorpdfstring{Q\textsubscript{f}}{like this}*) = encoder(Q) #[n, (*\textcolor{codegreen}{d\textsubscript{i}}*)]  

(*\texorpdfstring{R\textsubscript{f}}{like this}*) = encoder(R) #[n,(*\textcolor{codegreen}{d\textsubscript{i}}*)]
# project embedding [n, (*\textcolor{codegreen}{d\textsubscript{e}}*)]
(*\texorpdfstring{Q\textsubscript{e}}{like this}*) = l2_normalize(np.dot((*\texorpdfstring{Q\textsubscript{f}}{like this}*), (*\texorpdfstring{W\textsubscript{i}}{like this}*)), axis=1) 
(*\texorpdfstring{R\textsubscript{e}}{like this}*) = l2_normalize(np.dot((*\texorpdfstring{R\textsubscript{f}}{like this}*), (*\texorpdfstring{W\textsubscript{i}}{like this}*)), axis=1)

# scaled pairwise cosine similarities [n, n]
# t - learned temperature parameter
logits = np.dot((*\texorpdfstring{Q\textsubscript{e}}{like this}*), (*\texorpdfstring{R\textsubscript{e}}{like this}*).T) * np.exp(t) 
# symmetric loss function
(*\texorpdfstring{loss\textsubscript{Q}}{like this}*) = cross_entropy(logits, labels, axis=0) 
(*\texorpdfstring{loss\textsubscript{R}}{like this}*) = cross_entropy(logits, labels, axis=1)
loss = ((*\texorpdfstring{loss\textsubscript{Q}}{like this}*) + (*\texorpdfstring{loss\textsubscript{R}}{like this}*))/2
\end{lstlisting}
\caption{Numpy-like pseudo-code for the core of an implementation of our framework for Object Identification task.}
\label{fig:pseudocode}
\vspace{-5mm}
\end{figure}


\subsection{Object Identification}

Object identification is the second task in the ARMBench dataset. This task is concerned with precisely categorizing detected objects among a database of predefined classes. Unlike traditional classification approaches, which become computationally challenging with a large number of categories (over 190,000 in the Amazon warehouse scenario), we tackle this task as an image retrieval problem.

\subsubsection{Task Variants}

\looseness -1 In robotic manipulation within the context of the ARMBench dataset, the object identification task holds significance in pre- and post-object manipulation. In the pre-pick stage, object identification permits the retrieval of historically acquired objects or attributes for manipulation planning. While in the post-pick, the object's unique identifier is crucial for quality control and subsequent tasks. Thus, both pre- and post-pick images of the object could serve as query images, while the reference images in the container manifest act as gallery images. The challenge lies in accurately matching the query images to the gallery images.

Therefore, this task offers two variants: one reliant on pre-pick images (one for each pick) and the other incorporating post-pick images (two more for each pick). While the latter images present a greater challenge due to differing perspectives, object poses, and presentations, they enable the incorporation of multi-view data, thereby enhancing the overall retrieval performance.

\subsubsection{Overall Architecture}

Consistent with our object instance segmentation task strategy, we employ a backbone plus a task-specific head architecture as shown in Fig.~\ref{fig:framework}-center. For the two aforementioned task variants, the task-specific head comprises a projection head or adds a Multi-Layer Perceptron (MLP) for fusing pre- and post-pick images. A contrastive learning objective is applied to fine-tune the RoboLLM to optimize detection performance.

\subsubsection{Contrastive Learning Fine-tuning}

We argue that high-quality feature maps suffice for measuring the similarity between query and reference images using dot product calculations. We maintain computational efficiency by abandoning complex retrieval techniques such as re-ranking, which is crucial for real-time robotic applications. Thus, a contrastive objective is used in the fine-tuning stage to improve the quality of the generated feature maps. It aims to minimize the distance between the feature maps of positive pairs (the same class) and maximize the distance between negative pairs (different classes), thus enhancing the efficacy of our image retrieval approach.

\looseness -1 The backbone encoder transforms an input image or three images input \(Q\) into an output representation \({Q}_{f}\), which is further processed by a linear projection head to produce the final feature vector. In cases involving post-pick images as query images, their feature maps are concatenated before input into the MLP, providing a fused representation before feeding into the linear projection head. An \(L2\) normalization is applied to \({Q}_{f}\) to mitigate the risk of numerical instability during training. Specifically, given a batch of \(N\) (query image, reference image) pairs, our framework is trained to predict which of the \(N \times N\) possible (query image, reference image) pairings across a batch actually occurred. To do this, our framework maximizes the pairwise cosine similarity of the query and reference image feature maps of the \(N\) real pairs in the batch, while minimizing the cosine similarity of the feature maps of the \(N^2-N\) incorrect pairings. We optimize a symmetric cross-entropy loss over these similarity scores. In Figure~\ref{fig:pseudocode}, we include the pseudocode of the framework's core for the object identification task.

\subsubsection{Advantages and Rationale}

This approach presents multiple advantages. First, the scalability of the image retrieval method sidesteps the challenges tied to implementing a large-category linear classification layer. Second, the modular design allows task-specific heads to be easily interchanged, making the framework flexible and adaptable to different identification scenarios, whether involving single or multiple query images. Lastly, computational efficiency is assured through using a dot product similarity measure and the elimination of intricate retrieval techniques.

\vspace{-3mm}

\subsection{Defect Detection}

\vspace{-2mm}

The third task in the ARMBench dataset is defect detection, aimed at identifying defects resulting from specific robotic manipulation activities. This is a crucial task as it directly impacts the integrity of the workflow and the quality of the end product. The dataset includes two types of robot-induced defects: 1) multi-pick, where multiple objects are mistakenly picked and transferred from the source to the destination container; and 2) package-defect, indicating activities that result in the object's packaging opening or the object deconstructing into multiple parts.

Consistent with the design strategy for earlier tasks, we employ a backbone plus a task-specific-head architecture. We reuse the same backbone encoder employed in the previous tasks. Utilizing the same backbone across multiple tasks ensures a cohesive and streamlined architecture. As in the object identification task, we leverage segmentations to locate objects. Unlike the object identification task, which has many categories, this task comprises only three classes: two types of defects and a nominal type. Therefore, a classification head is sufficient to conduct the classification and is appended to the backbone encoder to make predictions across these three categories. We opt for a standard cross-entropy loss function for training, which is particularly suitable for categorical classification tasks.

\begin{table}[t]
\centering
\resizebox{0.49\textwidth}{!}{\begin{tabular}{@{}l|cc|cc|cc@{}}
\toprule
Task & \multicolumn{2}{c|}{Mixed Object Tote} & \multicolumn{2}{c|}{Zoomed Out Tote} & \multicolumn{2}{c}{Same Object Tote} \\ \midrule
Model & mAP50 & mAP75 & mAP50 & mAP75 & mAP50 & mAP75 \\ \midrule
ResNet50 + Mask RCNN * & 0.72 & 0.61 & 0.25 & 0.19 & 0.11 & 0.10 \\
RoboLLM & \textbf{0.82} & \textbf{0.67} & \textbf{0.57} & \textbf{0.45} & \textbf{0.15} & \textbf{0.13} \\ \bottomrule
\end{tabular}}
\caption{Mean Average Precision at IoU thresholds of 50 and 75 across the different segmentation task subsets. * indicates results obtained from~\cite{armbench}.}
\label{table:seg}
\vspace{-6mm}
\end{table}

\begin{figure}[t]
    \centering
    \includegraphics[width=7.3cm]{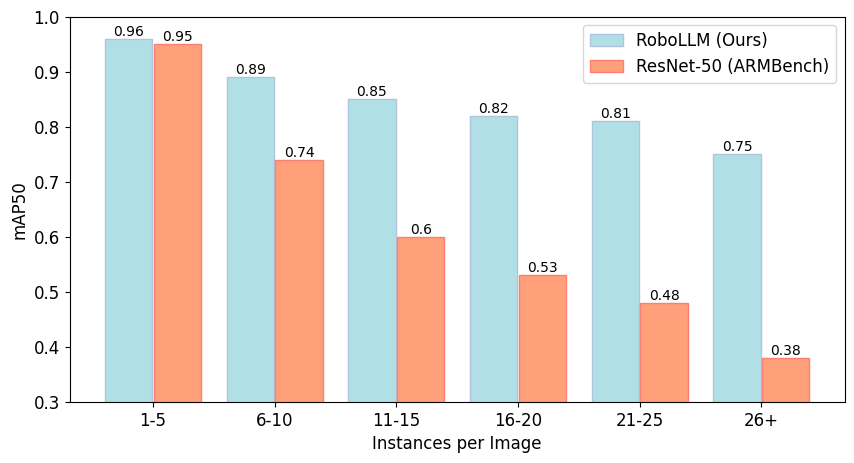}
    \caption{Number of object instances per image against Mean Average Precision at 50 on the Mixed-Object tote test-set.}
    \label{fig:instances}
    \vspace{-3mm}
\end{figure}

\begin{table}[t]
\centering
\resizebox{0.48\textwidth}{!}{\begin{tabular}{@{}lc|cc|cc|cc@{}}
\toprule
Model & Ref Set & \multicolumn{2}{c|}{Recall@1} & \multicolumn{2}{c|}{Recall@2} & \multicolumn{2}{c}{Recall@3} \\ \midrule
 &  & N=1 & N=3 & N=1 & N=3 & N=1 & N=3 \\ \midrule
ResNet50-RMAC & Container & 71.7 & 72.2 & 81.9 & 82.9 & 87.2 & 88.2 \\
DINO-ViT-S & Container & 77.2 & 79.5 & 87.3 & 89.4 & 91.6 & 93.5 \\
BEiT-3-Base* & Container & 83.7 & 84.5 & 83.8 & N/A & 84.5 & N/A \\
RoboLLM  & Container & \textbf{97.8} & \textbf{98.0} & \textbf{97.9} & \textbf{98.1} & \textbf{98.0} & \textbf{98.2} \\ \midrule
RoboLLM  & All refs & 74.6 & 78.2 & 82.6 & 85.7 & 85.3 & 89.10 \\ \bottomrule
\end{tabular}}
\caption{Results on object identification at varying recall@k. * indicates no ARMBench fine-tuning. N=1 uses one pre-pick image, while N=3 uses three images per pick. ``Ref set'' specifies if reference images are container-specific or from the entire dataset.}
\label{table:ir}
\vspace{-5mm}
\end{table}

\section{EXPERIMENTS\label{sec:experiments}}

\subsection{Experimental Setup}
Our experiments target the ARMBench dataset, focusing on three key robotic vision perception tasks: object instance segmentation, object identification, and defect detection. The experiments are conducted using two state-of-the-art Multi-Modal Learning Models (MLLMs) as the backbones of the proposed RoboLLM framework, namely BEiT-3 Base and BEiT-3 Large. Notably, the BEiT-3 Large model is employed exclusively for the defect detection task because the performance of the BEiT-3 Base model fails to meet the ideal task-specific requirements. We utilize the test sets of all tasks to report metrics. The backbones are initialized with pre-trained weights. A batch size of 64 is employed for all three tasks, with fine-tuning conducted over 50 epochs. An initial learning rate of \(2 \times 10^{-4}\) is set for all tasks. Early stopping is implemented if there is no improvement in the metrics on the validation set over a span of 5 epochs. For the object identification task, the Multi-Layer Perceptron (MLP) consists of two linear layers, with middle dimensions set to 3072 for the Base model and 4096 for the Large model. The projection head is implemented as a single linear layer, with its embedding dimension matching that of the encoder. 

\subsection{Object Segmentation}

\looseness -1 Object segmentation is a critical stage within the robotic vision pipeline, influencing robotic planning, grasp operations, and object identification. Poor object instance segmentation can introduce defects during robotic manipulation, which is highly undesirable. The ARMBench challenge offers three data subsets to evaluate image segmentation performance: 1) \textit{Mixed-Object-tote} serves as a general benchmark.  2) \textit{Zoomed-out-tote} assesses model generalization to new warehouse environments.  3) \textit{Same-Object} examines segmentation performance on tightly packed instances of identical objects.

We evaluate performance using mean-average-precision at Intersection-over-Union (IoU) thresholds of 0.5 (mAP50) and 0.75 (mAP75) across all subsets, as presented in Table \ref{table:seg}. Our RoboLLM framework significantly outperforms a ResNet-50 baseline across all tasks: 1) On the \textit{Mixed-Object-tote}, RoboLLM achieves 82\% and 67\% for mAP50 and mAP75, marking a 10\% and 6\% improvement over ResNet-50. 2) The performance for RoboLLM on the \textit{Zoomed-out-tote} is much better than it of ResNet-50, demonstrating superior generalization capabilities. 3) \textit{Same-Object} poses the most difficult segmentation challenge, yet BEiT-3 still improves mAP50 by 4\% over ResNet-50.

Furthermore, high variation in object instances between containers is common. To investigate model robustness under varying numbers of objects within an image, we report mAP50 in \textit{Mixed-Object-tote} for different numbers of object instances (Figure \ref{fig:instances}). While both RoboLLM and ResNet-50 perform similarly at fewer than five instances (approximately 95\% mAP), performance for ResNet-50 degrades significantly (to 38\%) when the number of instances exceeds 26. In contrast, RoboLLM demonstrates a modest decline, maintaining 75\% mAP for instances exceeding 26. This evidences that RoboLLM is robust for segmenting a large number of objects within an image.

\vspace{-1mm}
\subsection{Object Identification}
\vspace{-1mm}

Object identification requires categorizing segmented objects to facilitate robotic planning for subsequent maneuvers. We use Recall@k as the evaluation metric, and the resultant performances are reported in Table \ref{table:ir}.
Using the pre-trained BEiT-3-Base model as a strong baseline, we observe an uplift in recall@1 performance from 77.2\% to 83.7\%, outperforming the best previously reported result using DINO-ViT-S \cite{armbench}. Employing our RoboLLM framework, recall@1 is further improved to 97.8\% and 98.0\% when using only pre-pick images and both pre/post-pick images, respectively. This is a 21\% increase over the best prior result \cite{armbench}, effectively solving ARMBench's object identification task.

\subsubsection{Performance Under More Challenging Conditions}
We posit that this superlative performance could partially be credited to the specific challenge design, which restricts reference images to objects known to be present in the container. To evaluate RoboLLM in a more generalized scenario, we expand the set of reference images to include all unique objects within the dataset (190k+). As shown on the bottom line of Table~\ref{table:ir}, even under this more demanding setting, RoboLLM maintains an impressive 89.1\% at recall@3. This robust performance suggests that our framework can adeptly handle even more complex, large-scale retrieval problems than those posed by the ARMBench challenge. We also find that our framework benefits from including additional query images in this challenging setting. This corroborates our design choice to aggregate multiple query images into a single representation for retrieval, thereby enhancing the model's robustness and versatility.

\begin{table}[t]
\resizebox{0.49\textwidth}{!}{\begin{tabular}{@{}l|ccc|ccc|ccc@{}}
\toprule
Task & \multicolumn{3}{c|}{Multi-Pick} & \multicolumn{3}{c|}{Package Defect} & \multicolumn{3}{c}{Combined} \\ \midrule
Model & Precision & Recall & FPR & Precision & Recall & FPR & Precision & Recall & FPR \\ \midrule
ResNet50 * & - & 0.34 & 0.05 & - & 0.73 & 0.05 & - & 0.57 & 0.05 \\
RoboMLLM-B & \textbf{0.84} & 0.98 & \textbf{0.04} & \textbf{0.94} & 0.91 & 0.04 & \textbf{0.90} & 0.94 & \textbf{0.03} \\
RoboMLLM-L & 0.82 & \textbf{1.00} & \textbf{0.04} & 0.89 & \textbf{0.95} & \textbf{0.03} & 0.86 & \textbf{0.97} & \textbf{0.03} \\ \bottomrule
\end{tabular}}
\caption{Metrics for defect detection tasks with RoboMLLM B and L backbones. * denotes \cite{armbench} results. Combined metrics in right column. ARMBench ideal performance: $recall > 0.95$, $FPR < 0.01$.}
\label{table:defect}
\vspace{-7mm}
\end{table}

\subsection{Defect Detection}

The practical deployment of automated defect detection in commercial warehouse settings presents substantial challenges, given the high demand of the performance. To contextualize, the ARMBench challenge outlines ideal performance criteria, requiring a recall rate exceeding 0.95 and an FPR below 0.01. Our results for precision, recall, and FPR are reported in Table \ref{table:defect}. Our experiments demonstrate significant performance enhancements over a ResNet-50 baseline. Specifically, our RoboLLM with BEiT-Base achieves a combined recall rate of 94\% across both types of defect detection, marking a 37\% improvement over the baseline. Additionally, the combined FPR is reduced from 0.05 to 0.03. Despite these significant improvements, the system does not entirely meet the high recall and FPR criteria set forth by the ARMBench challenge.

To meet the desired performance set by ARMBench challenges, we conduct further experiments with a more powerful and larger model, BEiT-3 Large. Benefiting from our versatile yet robust framework, it is easy to adopt more powerful backbones as needed. Our results show that this model attains a combined recall rate of 97\%, surpassing the desired recall target. However, the FPR remains at 0.03, indicating room for future enhancements in effectiveness. Overall, our RoboLLM framework shows significant improvements over existing methods, while a gap in the FPR warrants further investigation in future work.


\vspace{-2mm}
\section{CONCLUSIONS}
In this paper, we introduce RoboLLM, an efficient and effective framework designed to establish a unified robotic vision pipeline aimed at the newly released Amazon ARMBench dataset~\cite{armbench}, utilizing a streamlined BEiT-3 as a Multi-Modal Large Language Model backbone encoder. We evaluate our framework on three distinct visual perception tasks: object segmentation, object identification, and defect detection, which are representative of large-scale real-world robotic vision problems. We show that RoboLLM significantly outperforms previous benchmarks across all three challenges. Notably, RoboLLM achieves this with inherent knowledge from pretraining on large-scale multimodal data and only a minimal task-specific head for each task, significantly increasing performance and mitigating the engineering challenges associated with each. Furthermore, the modular design of the proposed RoboLLM makes incorporating more powerful backbones easy, as well as task-specific modules, to push performance further in the future if required.

\clearpage

\bibliographystyle{IEEEtran}
\bibliography{References.bib}

\end{document}